\documentclass[twocolumn,twoside]{article}
\usepackage{cite}
\usepackage{epsfig,amsmath,amsfonts}
\usepackage{amsmath,amssymb,amsfonts,amsthm}
\usepackage{enumerate}
\usepackage{verbatim}
\usepackage{fancyhdr}
\usepackage{pxfonts}
\usepackage{graphicx}
\usepackage{framed}
\usepackage{mathrsfs}
\usepackage{epsfig}
\usepackage{leftidx}
\usepackage{appendix}
\usepackage{fancyhdr}
\usepackage{algorithm}
\usepackage{algpseudocode}

\theoremstyle{plain}
\newtheorem{theorem}{Theorem}[section]

\theoremstyle{definition}

\theoremstyle{remark}

\def\sech{\mathop{\operator@font sech}\nolimits}

\newtheorem*{thm*}{Theorem}

\newtheorem*{def*}{Definition}

\renewcommand{\epsilon}{\varepsilon}
\renewcommand{\phi}{\varphi}

\begin{document}

\title{\bf Bridging Physics-Informed Neural Networks with Reinforcement Learning: Hamilton-Jacobi-Bellman Proximal Policy Optimization (HJBPPO)}
\author{Amartya Mukherjee$^1$\thanks{Corresponding author.\hfil\break e-mail: a29mukhe@uwaterlooo.ca}~, Jun Liu$^1$
\\
\small{$^1$Department of Applied Mathematics, University of Waterloo, Waterloo, Ontario N2L 3G1, Canada}}

\date{}

\maketitle
\begin{abstract}
    This paper introduces the Hamilton-Jacobi-Bellman Proximal Policy Optimization (HJBPPO) algorithm into reinforcement learning. The Hamilton-Jacobi-Bellman (HJB) equation is used in control theory to evaluate the optimality of the value function. Our work combines the HJB equation with reinforcement learning in continuous state and action spaces to improve the training of the value network. We treat the value network as a Physics-Informed Neural Network (PINN) to solve for the HJB equation by computing its derivatives with respect to its inputs exactly. The Proximal Policy Optimization (PPO)-Clipped algorithm is improvised with this implementation as it uses a value network to compute the objective function for its policy network. The HJBPPO algorithm shows an improved performance compared to PPO on the MuJoCo environments.\\

\noindent
{\bf keywords:} Continuous-Time Reinforcement Learning, Physics-Informed Neural Networks, Proximal Policy Optimization, Hamilton-Jacobi-Bellman Equation
\end{abstract}



\section{Introduction}

In recent years, there has been a growing interest in Reinforcement Learning (RL) for continuous control problems. RL has shown promising results in environments with unknown dynamics through a balance of exploration in the environment and exploitation of the learned policies. Since the advent of REINFORCE with Baseline, the value network in RL algorithms has shown to be useful towards finding optimal policies as a critic network \cite{SutBar}. This value network continues to be used in state-of-the-art RL algorithms today.

In discrete-time RL, the value function estimates returns from a given state as a sum of the returns over time steps. This value function is obtained from solving the Bellman Optimality Equation. On the other hand, in continuous-time RL, the value function estimates returns from a given state as an integral over time. This value function is obtained by solving a partial differential equation (PDE) known as the Hamilton-Jacobi-Bellman (HJB) equation \cite{Munos1999_2}. Both equations are difficult to solve analytically and numerically, and therefore the RL agent must explore the environment and make successive estimations.

Currently existing algorithms in the RL literature such as Proximal Policy Optimization (PPO) aim to update the value function using the Bellman Optimality Equation so that it estimates the discrete-time returns for each state. However, we discovered that this value function, when trained on MuJoCo environments, does not show convergence towards the optimal value function as described by the HJB equation (see Figure \ref{fig:hjb_loss_PPO}). This shows that information is lost when the value function is trained using discrete time steps rather than continuous time.

The introduction of physics-informed neural networks (PINNs) by \cite{RAISSI2019686} has led to significant advancements in scientific machine learning. PINNs leverage auto-differentiation to compute derivatives of neural networks with respect to their inputs and model parameters exactly. This enables the laws of physics (described by ODEs or PDEs) governing the dataset of interest to act as a regularization term for the neural network. As a result, PINNs outperform regular neural networks on such datasets by taking advantage of the underlying physics of the data.

To the best of our knowledge, this paper is the first to examine the intersection between PINNs and RL. In order to force the convergence of the value function in PPO towards the solution of the HJB equation, we utilize PINNs to encode this PDE and bridge the information gap between returns computed over discrete time and continuous time. This allows our algorithm to utilize auto-differentiation to eliminate the error associated with gradient computation and discretization of time. We propose the Hamilton-Jacobi-Bellman Proximal Policy Optimization (HJBPPO) algorithm, which demonstrates superior performance in terms of higher rewards, faster convergence, and greater stability compared to PPO on MuJoCo environments, making it a significant improvement.

\section{Preliminaries}

Consider a controlled dynamical system modeled by the following equation:

\begin{equation}
    \dot x = f(x,u), \quad x(t_0)=x_0, 
\end{equation}
where $x(t)$ is the state and $u(t)$ is the control input. In control theory, the optimal value function $V^*(x)$ is useful towards finding a solution to control problems \cite{RLOFC}:

\begin{equation}
    \label{eq:value1}
    V^*(x)=\sup_{u}\int_{t_0}^{\infty}\gamma^t R(x(\tau;t_0,x_0,u(\cdot)),u(\tau))d\tau,
\end{equation}
where $R(x,a)$ is the reward function and $\gamma$ is the discount factor. The following theorem introduces a criteria for assessing the optimality of the value function [\cite{Liberzon2012}, \cite{Munos1999}].

\begin{theorem}
    \label{thm:1}
    A function $V(x)$ is the optimal value function if and only if:
    
    \begin{enumerate}
        \item $V\in C^1(\mathbb{R}^n)$ and $V$ satisfies the Hamilton-Jacobi-Bellman (HJB) Equation
        \begin{equation}
            V(x)\ln\gamma+\sup_{u\in U}\{R(x,u)+\nabla_xV^T(x)f(x,u)\}=0
        \end{equation}
        for all $x\in\mathbb{R}^n$.
        \item For all $x\in\mathbb{R}^n$, there exists a controller $u^*(\cdot)$ such that:
        \begin{align}
            &V(x)\ln\gamma+R(x,u^*(x))+\nabla_xV^T(x)f(x,u^*(x))\nonumber\\
            &=V(x)\ln\gamma+\sup_{\hat u\in U}\{R(x,\hat u)+\nabla_xV^T(x)f(x,\hat u)\}.\label{eq:optimalcontrol}
        \end{align}
    \end{enumerate}
\end{theorem}

Currently existing algorithms in RL do not focus on solving the HJB equation to maximize the total reward for each episode. For example, in PPO, the HJB equation does not seem to be satisfied when tested on MuJoCo environments. 

To show this, we define the HJB loss at each episode as the following:
\begin{align}
    &MSE_f\nonumber\\
    &=\frac{1}{T}\sum_{t=0}^{T-1}|V(x_t)\ln\gamma+R(x_t,a_t)+\nabla_xV^T(x_t)f(x_t,a_t)|^2\label{eq:HJBLoss1},
\end{align}
where $T$ is the number of timesteps in the episode, $x_t$ is the state of the environment at timestep $t$, and $a_t$ is the action taken at timestep $t$. $\nabla_xV^T(x_t)$ is computed exactly using auto-differentiation. We approximate $f(x_t,a_t)$ using finite differences:
\begin{align}
    \label{eq:HJBLoss2}
    &MSE_f=\frac{1}{T}\sum_{t=0}^{T-1}|V(x_t)\ln\gamma+R(x_t,a_t)\notag\\
    &\qquad\qquad\qquad\qquad+\nabla_xV^T(x_t)(\frac{x_{t+1}-x_t}{\Delta t})|^2,
\end{align}
where $\Delta x$ is the time step size used in the environment. We have plotted the HJB loss for each environment using PPO in Figure \ref{fig:hjb_loss_PPO}. The mean HJB loss for each environment takes extremely high values and does not show convergence in 6 out of 10 of the environments, thus showing that the value function does not converge to the optimal value function as shown by the HJB equation.

As a comparison, we have plotted the graphs for the value network loss in Figure \ref{fig:val_loss_PPO}. The Bellman optimality loss shows convergence in 8 out of the 10 environments. This shows that information is lost when we solve the solve the Bellman optimality equation for a discrete-time value function compared to continuous-time value function. It also shows that convergence of the value function does not necessarily lead to convergence in the HJB loss.

To solve this problem as shown in Figure \ref{fig:hjb_loss_PPO}, we treat the value network as a PINN and use gradient-based methods to reduce the HJB loss.

\section{Related Work}

The use of HJB equations for continuous RL has sparked interest in recent years among the RL community as well as the control theory community, and has led to promising works.
\cite{HJBDQL} introduced an alternate HJB equation for Q Networks and used it to derive a controller that is Lipschitz continuous in time. This algorithm has shown improved performance over Deep Deterministic Policy Gradient (DDPG) in three out of the four tested MuJoCo environments without the need for an actor network. 
\cite{DHJB} introduced a distributional HJB equation to train the FD-WGF Q-Learning algorithm. This models return distributions more accurately compared to Quantile Regression TD (QTD) for a particle-control task. Finite difference methods are used to solve this HJB equation numerically.
We intend to build up from these works by adding a policy network and by incorporating PINNs to solve the HJB equation. 
Furthermore, the authors mentioned the use of auto-differentiation for increased accuracy of the distributional HJB equation as a potential area for future research in their conclusion.
Our work relaxes the requirement for the controller to be Lipschitz, and it minimizes the computational error associated with finite difference methods.

The use of neural networks to solve the HJB equation has been an area of interest across multiple research projects.
\cite{DBLP:journals/corr/JiangCCT16} uses a structured Recurrent Neural Network to solve for the HJB equation and achieve optimal control for the Dubins car problem.
\cite{4267720} uses the Pineda architecture \cite{pineda1987generalization} to estimate partial derivatives of the value function with respect to its inputs. They used iterative least squares method to solve for the HJB equation. This algorithm shows convergence in several control problems without the need for an initial stable policy.
\cite{SIRIGNANO20181339} develops the DGM algorithm to solve PDEs. They use auto-differentiation to compute first order derivatives and Monte-Carlo methods to estimate higher order derivatives. This algorithm was used to solve the HJB equation to achieve optimal control for a stochastic PDE and achieved an error of 0.1\%. We intend to further advance from these works and use a PINN to solve for the HJB equation.

The use of a PINN to solve the HJB equation for the value network was done by \cite{HJBPINN} in an optimal feedback control problem setting. This mitigates the use of finite differences to compute derivatives of the value function. The paper achieves results similar to that of the true optimal control function in high dimensional problems.
We intend to build up from this work by using PINNs in a RL setting where the dynamics are unknown and exploration is needed.


\section{HJBPPO}

To our knowledge, our work is the first to combine the HJB equation with a currently existing RL algorithm, PPO. It is also the first to use a PINN to solve the HJB equation in a RL setting.

The PPO-Clipped algorithm is improvised with this implementation because it uses a value network to compute advantages used to update its policy network~\cite{PPO2017}. PPO is an actor-critic method that limits the update of the policy network to a small trust region at every iteration. This ensures that the objective function of the policy network is a good approximation of the true objective function and forces small updates to the value network as well. As a result, PPO shows state-of-the-art performance on deterministic RL environments by ensuring small and robust updates at every iteration.

A study by \cite{10.1287/moor.2017.0855} presented a convergence analysis for two time scaled stochastic approximation with controlled noise. In actor-critic methods, the parameter updates in neural networks using optimizers such as stochastic gradient descent or Adam can be seen as numerical solutions to a stochastic ODE.

As such, this work has been utilized by \cite{holzleitner2021convergence} to introduce a convergence analysis for actor-critic methods. Furthermore, these results have been used to show the asymptotic convergence of PPO and RUDDER \cite{arjona2019rudder}. The authors introduce model assumptions as well as loss function assumptions that need to be satisfied to ensure that parameters in PPO and RUDDER may converge to a local minimum in a neighborhood near their initial values. The study shows that the parameters of the policy network and value network in PPO may converge to a local minimum in a neighborhood near their initial values.

Another theoretical study by \cite{PPO_convergence} concludes that policy optimization methods including PPO shows a guaranteed convergence on LQR tasks through the use of gradient-based optimization by formulating it as a non-convex optimization problem. It also shows reliable performance on state-feedback control problems. The paper refers to advanced regularization techniques as a potential area for improving robustness. This further justifies the introduction of the HJB loss as a regularization term.

\subsection{PINNs for HJB equation}

Our work combines the HJB equation with reinforcement learning in continuous state and action spaces to improve the training of the value network. On a stochastic system with infinite time horizon, the HJB equation is a second order elliptic equation \cite{ChangFR}. A theoretical study by \cite{PINN_convergence} shows that PINNs converge uniformly to the solution of second order linear elliptic equations, thus justifying the use of PINNs to solve the HJB equation.

We treat the value network as a PINN to solve for the HJB equation by computing its derivatives respect to its inputs exactly.

Note that the term 
$$\sup_{\hat u\in U}\{R(x,\hat u)+\nabla_xV^T(x)f(x,\hat u)\}$$ 
in the HJB equation cannot be determined without exploration of the agent in its environment. From Theorem \ref{thm:1}, the optimal policy $\pi^*(a|x)$ and the optimal controller $u^*(x)=\text{argmax}_a\pi^*(a|x)$ satisfies equation (\ref{eq:optimalcontrol}). The optimal policy is modeled by the policy network $\pi_\theta$ parameterized by $\theta$ and the optimal controller can be approximated using $u(x)=\text{argmax}_a\pi_\theta(a|x)$. As a result, we can use the following approximation:
\begin{align}	&V(x)\ln\gamma+R(x,u(x))+\nabla_xV^T(x)f(x,u(x))\nonumber\\
    &\approx V(x)\ln\gamma+\sup_{\hat u\in U}\{R(x,\hat u)+\nabla_xV^T(x)f(x,\hat u)\}.
    \label{eq:approximation}
\end{align}
This, as a result, justifies the use of equation \ref{eq:HJBLoss2} as the HJB loss used to update the value function at each episode.

The loss function is computed as 
$$J(\phi)=0.5MSE_u+\lambda_{HJB}MSE_f,$$ 
where $MSE_f$ is defined in equation (\ref{eq:HJBLoss2}) and $MSE_u$ is the standard loss function for the value network used in PPO, and is used to improve the discrete time estimate of returns for the value function:
\begin{equation}
    \label{eq:ValueLoss}
    MSE_u=\frac{1}{T}\sum_{t=0}^{T-1}|V(x_t)-(R(x_t,a_t)+\gamma V(x_{t+1}))|^2,
\end{equation}

where $\{x_t\}_{t=1}^{T}$ is a batch of states explored in a single episode, and $\{a_t=\text{argmax}_a\pi_\theta(a|x_t)\}_{t=1}^{T}$ is a batch of actions executed at time step $t$ following the policy $\pi_\theta$. The hyperparameter $\lambda_{HJB}$ is determined based on the magnitude of the HJB loss curves compared to the Bellman optimality loss curves so that both loss functions are given similar weight.

\subsection{Algorithm}

The HJBPPO algorithm is provided in Algorithm \ref{alg:HJBPPO}. The policy update and the minimization of $MSE_u$ for the value network is identical to PPO. In order to satisfy $V(x)\in C^1(\mathbb{R}^n)$ as stated in Theorem \ref{thm:1}, we use the infinitely differentiable $tanh$ activation function for the value network.

\begin{algorithm}[tb]
    \caption{HJBPPO}
    \label{alg:HJBPPO}
    \begin{algorithmic}[1]
        \State Initiate policy network parameter $\theta$ and value network parameter $\phi$
        \For{$iteration=1,2,...$}
        \State Run the policy $\pi_\theta$ in the environment for $T$ timesteps and observe samples $\{(s_t,a_t,R_t,s_{t+1})\}_{t=1}^{T}$.
        \State Compute the advantage $A_t$
        \State Compute $r_t(\theta)=\frac{\pi_\theta(a_t|s_t)}{\pi_{\theta_{\text{old}}}(a_t|s_t)}$
        \State Compute the objective function of the policy network: $$L(\theta)=\frac{1}{T}\sum_{t=0}^{T-1}\min[r_t(\theta)A_t,\text{clip}(r_t(\theta),1-\epsilon,1+\epsilon)A_t]$$
        \State Update $\theta\leftarrow\theta+\alpha_1\nabla_\theta L(\theta)$
        \State Compute the value network loss as: $J(\phi)=0.5MSE_u+\lambda_{HJB}MSE_f$ described in equations (\ref{eq:ValueLoss}) and (\ref{eq:HJBLoss2})
        \State Update $\phi\leftarrow\phi-\alpha_2\nabla_\phi J(\phi)$
        \EndFor
    \end{algorithmic}
\end{algorithm}

Lines 3--7 in the algorithm are identical to the PPO algorithm. The advantage term $A_t$ is computed as: $A_t=\sum_{n=t}^{T-1}(\gamma\lambda)^{n-t}\delta_{n}$, where $\delta_t=R_t+\gamma V_{\phi}(s_{t+1})-V_{\phi}(s_t)$, $\gamma$ is the discount factor, and $\lambda$ is the generalized advantage estimation (GAE) parameter. $\alpha_1$ and $\alpha_2$ are learning rates for the policy network optimizer and value network optimizer respectively. $\epsilon$ is the clipping parameter.

Lines 8--9 in the algorithm are our modifications to the PPO algorithm. We treat the value network as a PINN and add a $MSE_f$ term into its loss function. This way, the HJB equation is used as a regularization term for the value network.

We will compare the performance of HJBPPO to PPO on the MuJoCo environments for rewards as well as the HJB loss.

\section{Results}

\begin{figure}[t]
    \centering
    \includegraphics[width=\linewidth]{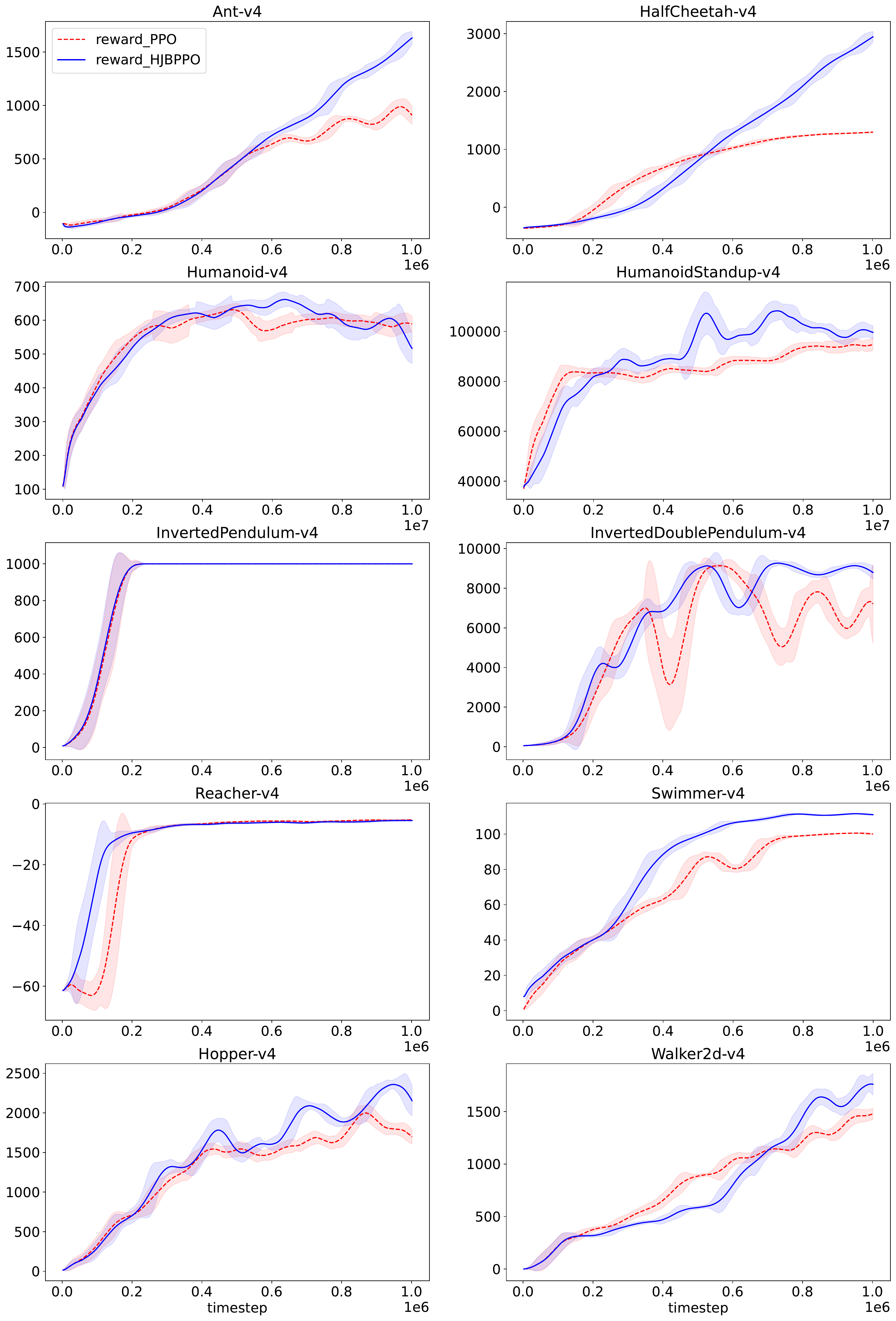}
    \caption{Comparison of learning curves for PPO (Red, Dashed) compared to HJBPPO (Blue, Smooth)}
    \label{fig:reward}
\end{figure}

\begin{figure}[t]
    \centering
    \includegraphics[width=\linewidth]{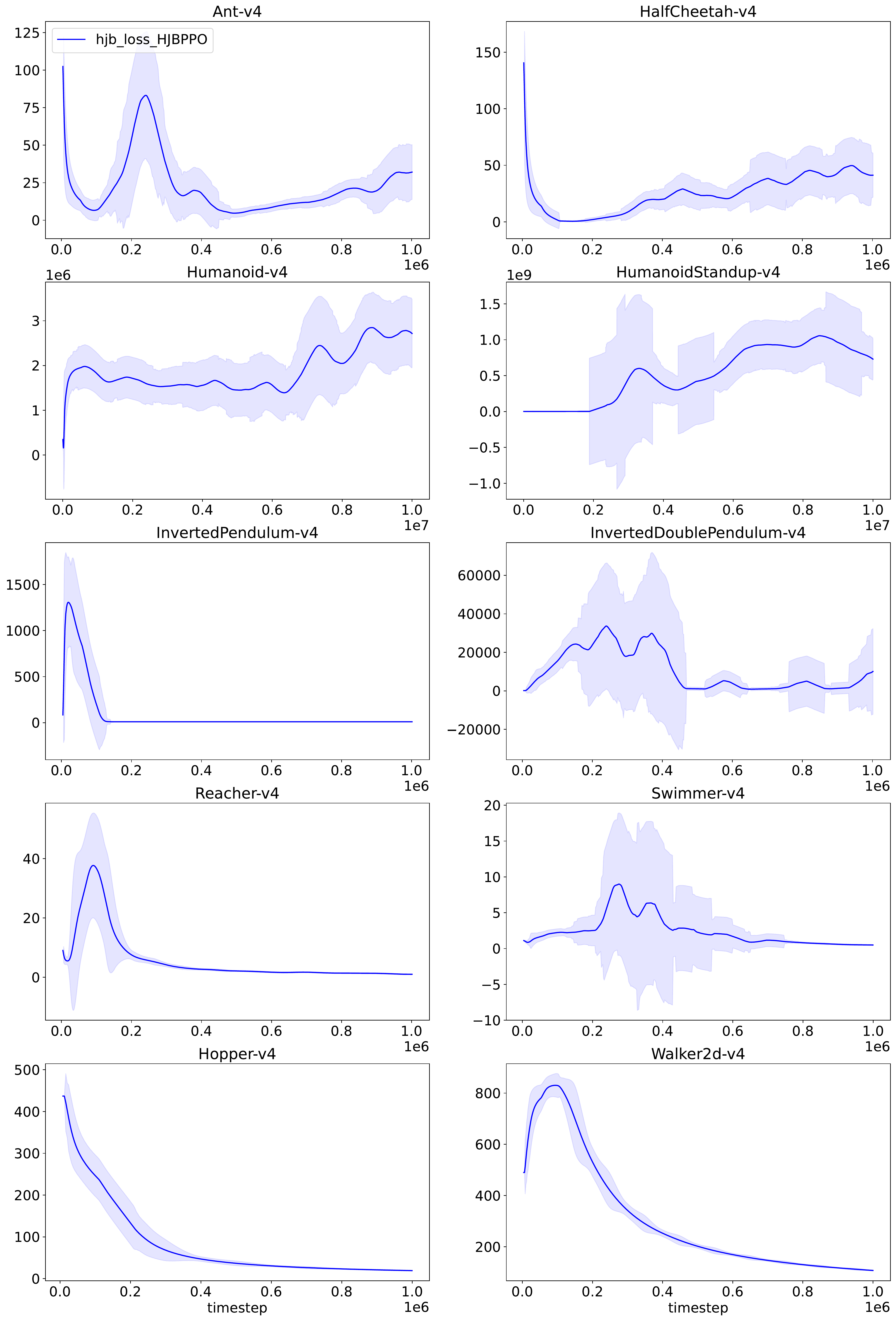}
    \caption{HJB loss curves for HJBPPO on MuJoCo environments}
    \label{fig:hjb_loss_HJBPPO}
\end{figure}

\begin{figure}[t]
    \centering
    \includegraphics[width=\linewidth]{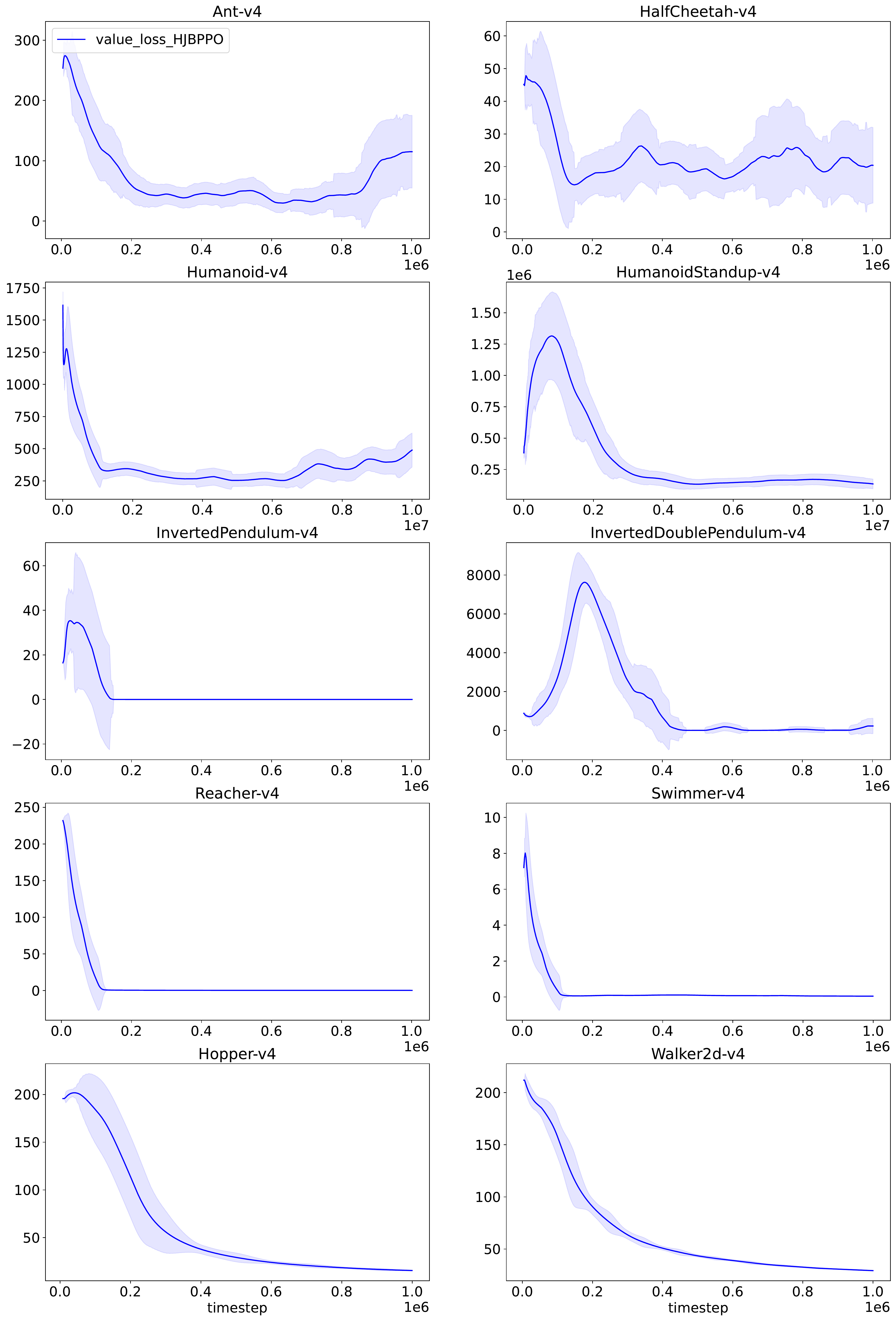}
    \caption{Bellman optimality loss curves for HJBPPO on MuJoCo environments}
    \label{fig:val_loss_HJBPPO}
\end{figure}

\begin{figure}[t]
    \centering
    \includegraphics[width=\linewidth]{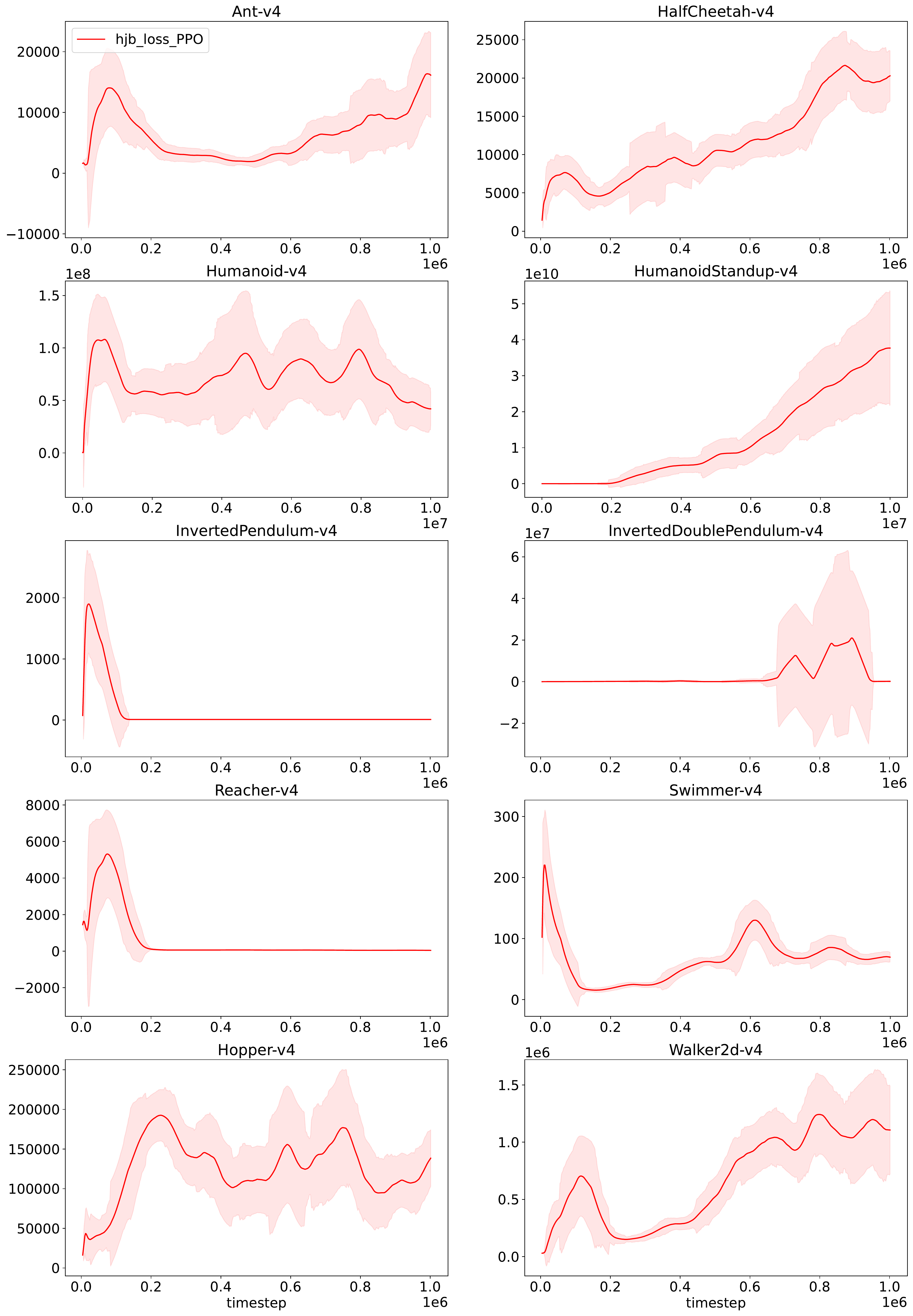}
    \caption{HJB loss curves for PPO on MuJoCo environments}
    \label{fig:hjb_loss_PPO}
\end{figure}

\begin{figure}[t]
    \centering
    \includegraphics[width=\linewidth]{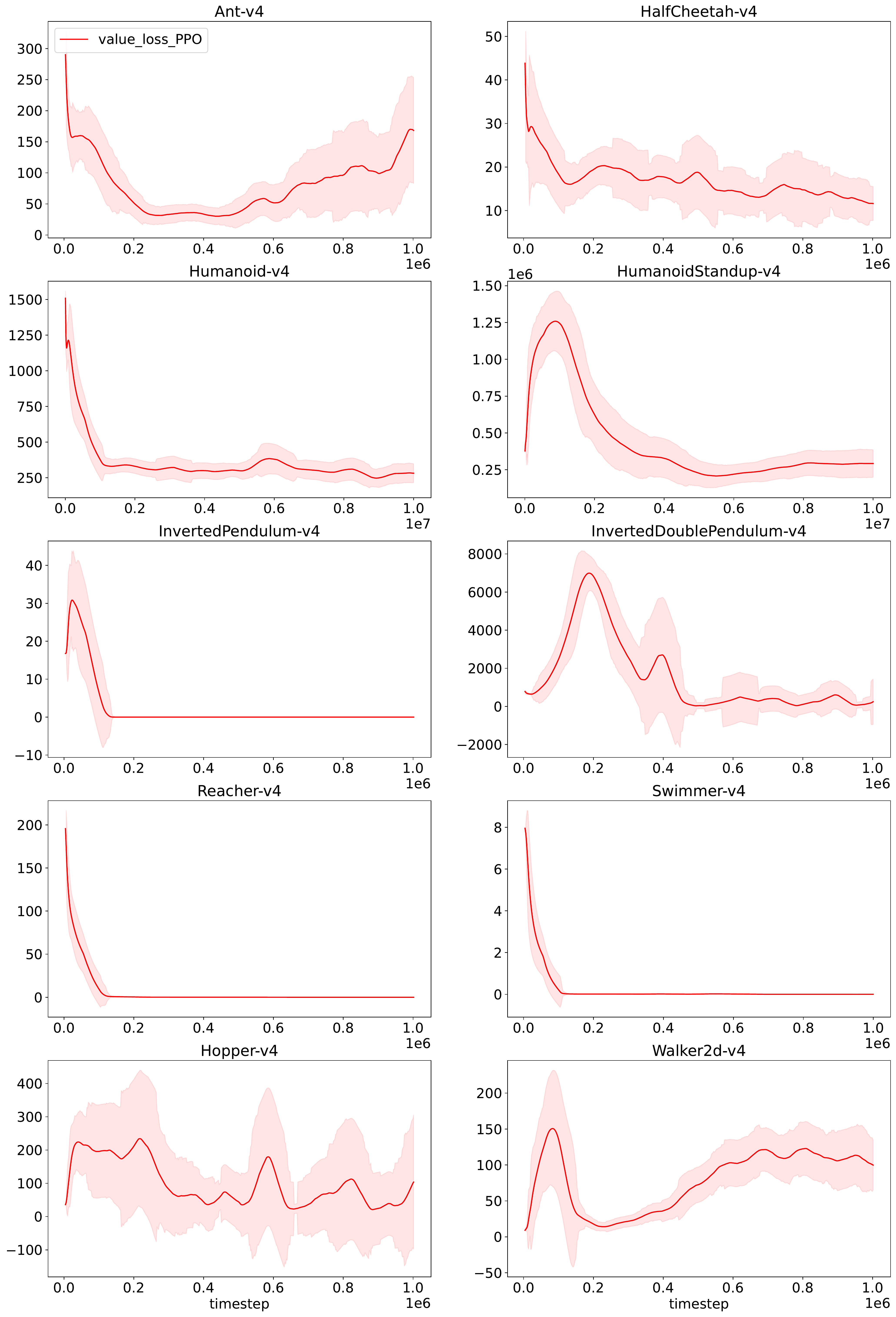}
    \caption{Bellman optimality loss curves for PPO on MuJoCo environments}
    \label{fig:val_loss_PPO}
\end{figure}

\begin{figure}[t]
    \centering
    \includegraphics[width=\linewidth]{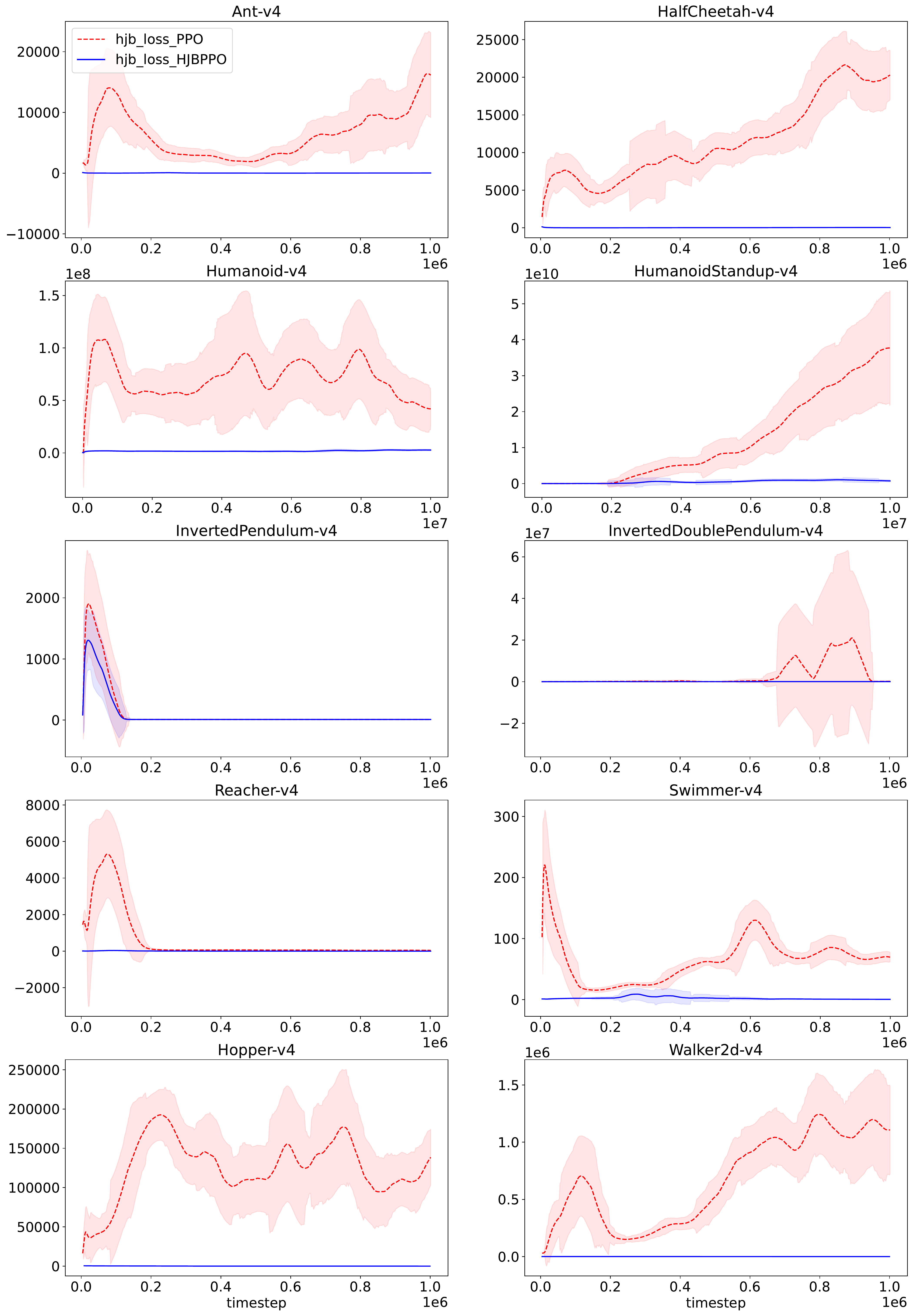}
    \caption{Comparison of HJB loss curves for PPO (Red, Dashed) compared to HJBPPO (Blue, Smooth)}
    \label{fig:hjb_loss}
\end{figure}

\begin{figure}
    \centering
    \includegraphics[width=\linewidth]{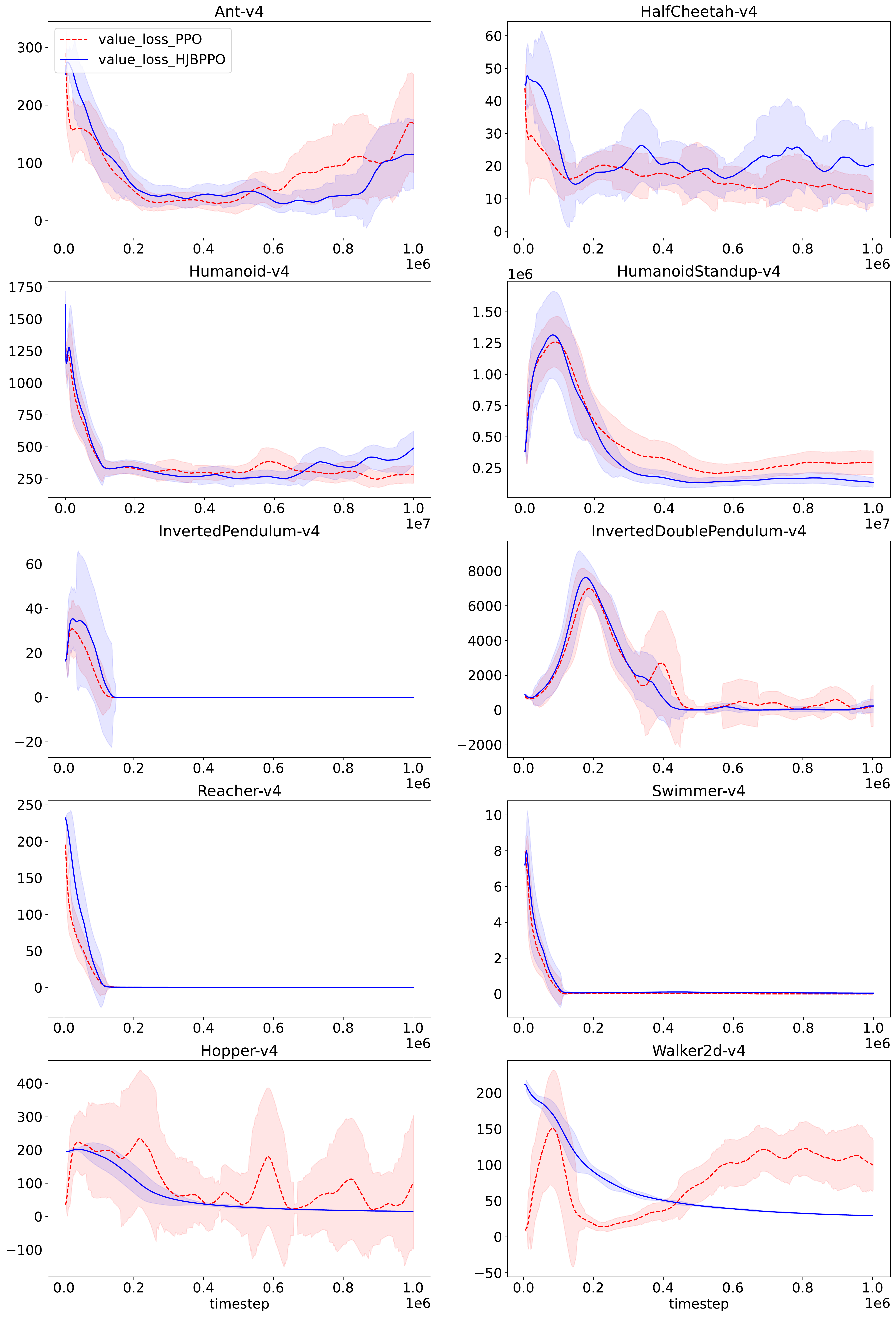}
    \caption{Comparison of Bellman optimality loss curves for PPO (Red, Dashed) compared to HJBPPO (Blue, Smooth)}
    \label{fig:val_loss}
\end{figure}

\subsection{Training}

The HJBPPO algorithm was implemented by modifying the code for PPO in the Stable Baselines 3 library by \cite{stable-baselines3}. To ensure the reproducibility of our results, we have posted our code at (Github repository redacted, code provided in supplementary material) and we have provided our hyperparameters in Tables \ref{tab:hparams} and \ref{tab:hparams_lambda} in Appendix \ref{sec:app:hyperparams}.

The code was run on the Béluga cluster in Compute Canada. The cluster provided the MuJoCo environments for training. Training each algorithm over 1 million time steps took seven hours, and training over 10 million time steps took three days. The multiprocessing library from python was used to train each algorithm over multiple environments at the same time.

\subsection{Reward Curves}

The reward graphs have been plotted in Figure \ref{fig:reward}, comparing HJBPPO to PPO on all the MuJoCo environments over a million time steps or ten million time steps. The line shows the total reward, averaged over 50 consecutive episodes, and the shaded area indicated the standard deviation of the total reward over 50 consecutive episodes. HJBPPO shows a significant improvement in Ant-v4 and HalfCheetah-v4. It shows faster convergence and stability in Reacher-v4, Swimmer-v4, and InvertedDoublePendulum-v4. And it shows a slight improvement in HumanoidStandup-v4, Hopper-v4, and Walker2d-v4. For the two remaining environments (Humanoid-v4 and InvertedPendulum-v4), it shows equal performance to PPO.

As a result, the graphs show that incorporating the continuous-time HJB equation into the PPO algorithm to train the value function leads to an improved learning curve for the agent. This is because HJBPPO uses a PINN to exploit the physics in the environment. It uses finite differences to approximate the underlying governing equation $f(x,u)$ of the environment and uses auto-differentiation to solve the HJB equation to achieve optimal control.

\subsection{HJB Loss Curves}

The HJB loss for each environment has been plotted in Figure \ref{fig:hjb_loss_HJBPPO}. HJBPPO shows a significant decrease in the HJB loss compared to PPO. And the HJB loss shows convergence in 8 out of the 10 environments, including Ant-v4 and HalfCheetah-v4 where it performs significantly better than PPO in terms of rewards, thus showing that the value function converges to the optimal value function as shown in the HJB equation.

The HJB loss does not converge for Humanoid-v4 and HumanoidStandup-v4, even though the HJB loss for these environments is significantly lower than that as shown in Figure \ref{fig:hjb_loss_PPO}. In both of these environments, the reward curve for HJBPPO shows similar performance to PPO. This shows that HJBPPO does not show significantly improved performance compared to PPO in general if the HJB loss does not show convergence.

\subsection{Bellman Optimality Loss Curves}

The Bellman optimality loss for each environment has been plotted in Figure \ref{fig:val_loss_HJBPPO}. The value network shows convergence in every environment. This shows that convergence of the value function in the continuous-time HJB equation also improves its convergence in the discrete-time Bellman optimality equation, while the converse may not necessarily be true.

\subsection{Comparison with PPO}

For comparison, we have posted the HJB loss and Bellman optimality loss curved for PPO in Figures \ref{fig:hjb_loss_PPO} and \ref{fig:val_loss_PPO} below. A notable difference is that the HJB loss in HJBPPO takes significantly lower valued compared to PPO. This is because the HJB loss is actively being minimized during the training of HJBPPO. To make the difference clearer, we plotted the loss curves on the same graphs in Figure \ref{fig:hjb_loss}. It is clear that the HJB loss for HJBPPO takes extremely small values in comparison with PPO.

As shown in Figure \ref{fig:hjb_loss_PPO}, the HJB loss in PPO shows convergence in only 4 out of the 10 environments; InvertedPendulum-v4, InvertedDoublePendulum-v4, Reacher-v4, and Swimmer-v4. For the remaining environments, the HJB loss shows an overall increasing trend.

For the environments where the HJB loss converges for PPO, it also shows convergence for HJBPPO as shown in Figure \ref{fig:hjb_loss_HJBPPO}. While HJBPPO does not show convergence in HJB loss for HumanoidStandup-v4, the loss curve is an improvement compared to PPO, where the loss curve shows a significant increasing trend.

Figure \ref{fig:val_loss_PPO} shows convergence in Bellman optimality loss for 8 out of the 10 MuJoCo environments using PPO. The convergence in the Bellman optimality loss is achieved by PPO by training the value function to solve for the Bellman optimality equation. However, despite the choice of this loss function, in Ant-v4 and Walker2d-v4, the Bellman optimality loss does not show convergence, and instead shows an increasing trend for large time steps.

This issue is solved in HJBPPO as shown in Figure \ref{fig:val_loss_HJBPPO}. The Bellman optimality loss shows an overall decreasing trend in all environments including Ant-v4 and Walker2d-v4. To make the difference clearer, we plotted the Bellman optimality loss curves on the same graphs in Figure \ref{fig:val_loss}. The Bellman optimality loss curves for HJBPPO show equal performance in general compared to PPO with better convergence in Ant-v4, HumanoidStandup-v4, InvertedDoublePendulum-v4, Hopper-v4, and Walker2d-v4.

In summary, HJBPPO shows an improved performance compared to PPO. It shows improvement in the rewards curves, the HJB loss curves, and the Bellman optimality loss curves. This is due to the fact that HJBPPO incorporates an HJB loss regularization term and uses works from optimal control to improve the learning of the value function, and thus, improve the convergence of the algorithm.

\section{Conclusion}

In this paper, we have introduced the HJBPPO algorithm that improvises the PPO algorithm to solve the HJB equation. This paper is the first of its kind to combine PINNs with RL. We treat the value function as a PINN to solve the HJB equation in an RL setting. The HJBPPO algorithm shows an overall improvement in performance compared to PPO due to its ability to exploit the physics of the environment as well as optimal control to improve the learning curve of the agent. This paper also shows that convergence of the value function in the continuous-time HJB equation also improves its convergence in the discrete-time Bellman optimality equation.

\section{Future Research}

Despite showing an overall improvement in the reward curves, the HJBPPO leaves room for improved RL algorithms using PINNs. 

A limitation of the HJBPPO algorithm as shown in figure \ref{fig:hjb_loss_HJBPPO} is that the HJB loss does not always show convergence in the environments albeit showing a significant improvement compared to PPO. A potential area of further research could involve new optimization methods for PINNs that show improved convergence of the HJB loss.

The loss function $MSE_f$ using in training the value network was derived as a result of the approximation used in equation \ref{eq:approximation}. So this does not guarantee convergence of the policy network towards the optimal policy such that $u(x)=\sup_{\hat u\in U}\{R(x,\hat u)+\nabla_xV^T(x)f(x,\hat u)\}$ where the controller $u(x)$ is derived from the policy $\pi_\theta(a|x)$. \cite{holzleitner2021convergence} proves the convergence of the policy network parameters in PPO to a local optimum but it does not guarantee global convergence. Thus, a potential area of further research could involve alternate choices of HJB loss functions for the value network that relaxes this approximation.

In this paper, we have explored and compared two deterministic RL algorithms - HJBPPO and PPO. It will be interesting to see how this algorithm can be extended to a stochastic setting. In the SAC paper, \cite{SAC} introduces an entropy-regularized stochastic policy that is less likely to overfit or stick to a local optima. As a consequence of the approximation used in equation (\ref{eq:approximation}), the HJBPPO algorithm also poses a risk that the HJB loss of the value function could lead it to overfit to a suboptimal policy. This risk could be lessened by introducing an alternate HJB equation that facilitates exploration and incorporates entropy maximization. As a result, improvising the SAC algorithm with this HJB equation using PINNs is a potential area for further exploration.

In the MuJoCo environments, the HJBPPO algorithm showed an improvement compared to PPO. But this is due to the fact that $f(x,u)$ could be estimated through finite differences, thus allowing for the physics of the environment to be exploited. The environments give all the details of the state needed to choose an action. One limitation of HJBPPO is that it may not perform well in partially observable environments because the estimate of $f(x,u)$ may be inaccurate. Deep Transformer Q Network (DTQN) was introduced by \cite{DTQN} and achieves state-of-the-art results in many partially observable environments. A potential area for further research may be the introduction of an alternate HJB equation that facilitates partial observability. The DTQN algorithm may be improvised by incorporating this HJB equation using PINNs.

Additionally, finite difference approximations become less accurate in environments with high dimensions \cite{SIRIGNANO20181339}. This makes the HJB loss less reliable in environments such as Humanoid-v4 and HumanoidStandup-v4 where the state is a 376-dimensional vector. The finite difference approximation does not compute $f(x,u)$ exactly because the environment uses semi-implicit Euler integration steps rather than Euler's method \cite{6386025}. Thus, a potential area for future research could be combining HJBPPO with model-based RL so that $f(x,u)$ can be estimated with a smaller error.

\section*{Acknowledgments}
The authors would like to thank Pascal Poupart, Ashish Gaurav, and Yanting Miao from the Department of Computer Science, University of Waterloo, for providing us with feedback for this paper.

\newpage
\appendix
\onecolumn
	
\section{Hyperparameters}\label{sec:app:hyperparams}

\begin{table}[H]
    \centering
    \begin{tabular}{ c|c } 
        Hyperparameter & Value\\
        \hline
        Horizon (T) & 2048 \\ 
        Adam stepsize & 3e-04 \\ 
        Num.  epochs & 10 \\ 
        Minibatch size & 64 \\ 
        Discount ($\gamma$) & 0.99 \\ 
        GAE parameter ($\lambda$) & 0.95 \\ 
    \end{tabular}
    \caption{HJBPPO hyperparameters}
    \label{tab:hparams}
\end{table}

\begin{table}[H]
    \centering
    \begin{tabular}{c|c}
        Environment & Value\\
        \hline
        Ant-v4 & 0.1 \\ 
        HalfCheetah-v4 & 0.1 \\ 
        Humanoid-v4 & 1e-04 \\ 
        HumanoidStandup-v4 & 1.0 \\ 
        InvertedPendulum-v4 & 1e-04 \\ 
        InvertedDoublePendulum-v4 & 1e-03 \\ 
        Reacher-v4 & 1.0 \\ 
        Swimmer-v4 & 1e-04 \\ 
        Hopper-v4 & 0.1 \\ 
        Walker2d-v4 & 0.1 \\ 
    \end{tabular}
    \caption{$\lambda_{HJB}$ hyperparameter for each environment}
    \label{tab:hparams_lambda}
\end{table}


\end{document}